\documentclass{ifacconf}

\usepackage{graphicx}      
\usepackage{natbib}        
\usepackage{subcaption}
\usepackage{caption}
\usepackage{amsmath}
\begin{document}
\begin{frontmatter}

\title{Abdominal Multi-Organ Segmentation Based on Feature Pyramid Network and Spatial Recurrent Neural Network} 

\thanks[footnoteinfo]{© 2023 the authors. This work has been accepted to IFAC for publication under a Creative Commons Licence CC-BY-NC-ND}

\author{Yuhan Song,} 
\author{Armagan Elibol,} 
\author{Nak Young Chong}

\address{School of Information Science, Japan Advanced Institute of Science and Technology, Japan 923-1292 (e-mail: s2010098@jaist.ac.jp, aelibol@jaist.ac.jp, nakyoung@jaist.ac.jp)}

\begin{abstract}                
As recent advances in AI are causing the decline of conventional diagnostic methods, the realization of end-to-end diagnosis is fast approaching. Ultrasound image segmentation is an important step in the diagnostic process. An accurate and robust segmentation model accelerates the process and reduces the burden of sonographers. In contrast to previous research, we take two inherent features of ultrasound images into consideration: (1) different organs and tissues vary in spatial sizes, (2) the anatomical structures inside human body form a relatively constant spatial relationship. Based on those two ideas, we propose a new image segmentation model combining Feature Pyramid Network (FPN) and Spatial Recurrent Neural Network (SRNN). We discuss why we use FPN to extract anatomical structures of different scales and how SRNN is implemented to extract the spatial context features in abdominal ultrasound images.
\end{abstract}


\end{frontmatter}

\section{Introduction}

\begin{figure}
	\centering
	\includegraphics[width=0.7\linewidth]{"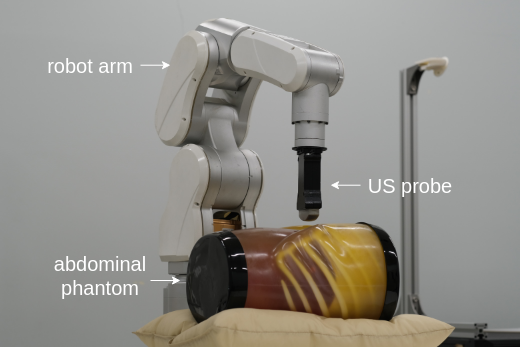"}
	\caption{Our robot-assisted ultrasound imaging system}
	\label{fig:russ}
\end{figure}

As many countries face the challenges of population aging with healthcare staff shortages, the demand for remote patient monitoring drives the development of AI-assisted diagnosis. In clinical practice, ultrasound imaging is one of the most common imaging modalities due to its effectiveness, non-invasive, and non-radiation nature. 
Medical ultrasound imaging requires an accurate delineation or segmentation of different anatomical structures for various purposes , like guiding the interventions. However, compared with other modalities, it is relatively harder to process because of low contrast, acoustic shadows, and speckles, to name a few (\cite{8614204}). It can be challenging even for experienced sonographers to detect the exact contour of tissues and organs, not to mention that it usually takes years of study and practice to train a qualified sonographer. Therefore, an automated and robust ultrasound image segmentation method is expected to help with locating and measuring important clinical information. Along the lines, we are developing a control algorithm for the robot arm to perform automatic ultrasound scans (see Fig. \ref{fig:russ}). Because this system is expected to operate automatically without human intervention, an evaluation metric for the robot's movement is necessary. To this end, a segmentation algorithm needs to be incorporated into the robot trajectory control system.

Traditional ultrasound image segmentation methods focus on the detection of textures and boundaries based on morphological or statistical methods. \cite{MISHRA2003967} proposed an active contour solution using low pass filters and morphological operations to make a prediction of the cardiac contour. \cite{Mignotte2001} developed a boundary estimation algorithm based on a Bayesian framework, where the estimation problem was formulated as an optimization algorithm to maximize the posterior possibility of being a boundary. Previously \cite{MIGNOTTE2001265} used statistical external energy in a discrete activate contour for the segmentation of short-axis parasternal images, in which a shifted Rayleigh distribution was used to model gray-level statistics. \cite{BOUKERROUI2003779} also proposed a Bayesian framework to conduct robust and adaptive region segmentation, taking the local class mean with a slow spatial variation into consideration to compensate for the nonuniformity of ultrasound echo signals.

Ultrasound image segmentation is time-consuming and prone to irregular anatomical structure shapes, and requires manual initialization operations. Compared with morphological and statistical methods, convolutional neural network (CNN) based solutions are powerful and flexible because of their strong nonlinear learning ability. 
\cite{7822557} conducted coarse and fine lymph node segmentation based on two series-connected fully convolutional networks. \cite{crossorgan} proposed a modified deep residual U-Net model to predict the contour of abdominal organs and tissues. They train their model initially on a tendon dataset, then fine-tune it on a breast tumor dataset. After getting a pre-trained model, they adapt the model to detect different anatomical structures using transfer learning. \cite{malepelvic} proposed a male pelvic multi-organ segmentation method on transrectal ultrasound images. In their research, a fully convolutional one-state (FCOS) object detector originally designed for generalized object detection is adapted for ultrasound image segmentation.

In the context of abdominal ultrasound image segmentation, most of the existing methods are targeted at specific organs or anomalies. \cite{CHEN2022106712} designed a multi-scale and deep-supervised CNN architecture for kidney image segmentation. They implemented a multi-scale input pyramid structure to capture features at different scales, and developed a multi-output supervision module to enable the network to predict segmentation results from multi-scales. \cite{10.1371/journal.pone.0219369} developed a detection algorithm for pulmonary nodules based on deep three-dimensional CNNs and ensemble learning. However, the importance of multi-organ segmentation is still ignored. On one hand, the segmentation result can be used as a guide for the remote ultrasound scan system, which is the cornerstone of realizing an automatic remote diagnostic system. An automatic diagnostic system will reduce burdens on sonographers, enabling them to concentrate on the analysis of pathology. On the other hand, abdominal organ segmentation can also provide information on specific organs and tissues, which can be used to assist in the diagnosis of certain diseases. Therefore, in this paper, an abdominal multi-organ segmentation method is proposed. Our contribution can be listed as follows: (a) We proposed a multi-organ segmentation method based on an FPN structure; (b) We combined the FPN model with an SRNN module, which helps improve the performance significantly.

\section{Related Methods based on CNNs}

\subsection{ResNet}

We use ResNet (\cite{He2016DeepRL}) as the feature extractor backbone, a deep residual learning framework to solve the degradation problem of deep networks. A common issue in deep learning is that deeper neural networks are harder to train. With the layers going deeper, accuracy would drop rapidly. In other words, appending more layers to a suitably deep model will increase the training error. \cite{He2016DeepRL} addressed the degradation problem by reformulating some of the layers as learning residual functions with reference to the layer inputs, instead of learning unreferenced functions. 

\begin{figure}
	\centering
	\includegraphics[width=0.7\linewidth]{"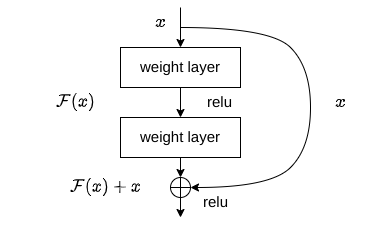"}
	\caption{Shortcut connection in ResNet}
	\label{fig:shortcut-connection}
\end{figure}

ResNet makes use of feedforward networks with ``shortcut connections'', which makes the network easier to optimize and able to gain extra accuracy from considerably increased depth. In the aforementioned paper, the authors let the shortcut connections simply perform identity mapping and produce the sum of the output from original layers and the lateral layers as illustrated in Fig. \ref{fig:shortcut-connection}, where $x$ represents the input, $F(x)$ the abstract representation of the residual block, and relu the activation function. $x$ is passed directly to the output and called ``identity shortcut connection''. Inserting the shortcut connection to the plain backbone, they managed to train models with over 1000 layers. ResNet's capacity of extracting deep features made it possible in our work to combine both semantic and abstract information for analyzing ultrasound images.

\subsection{Feature Pyramid Network}
\begin{figure}

	\begin{subfigure}{0.425\linewidth}
	\includegraphics[width=\linewidth,height=0.75\linewidth]{"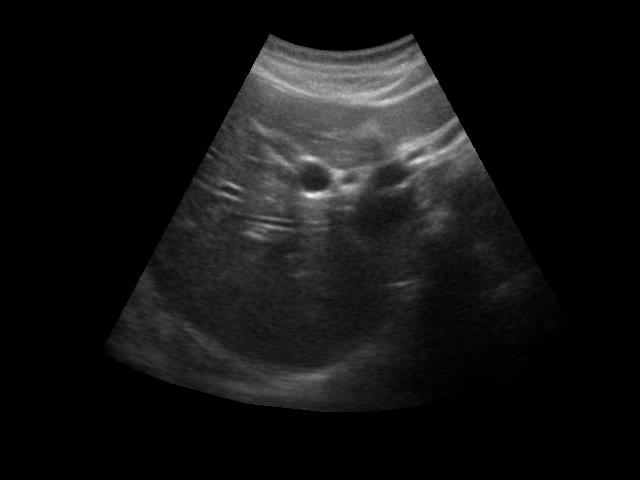"}
	\caption{Ultrasound image}\label{fig:ultimage}
	\end{subfigure}
	\hfill
	\begin{subfigure}{0.425\linewidth}
	\includegraphics[width=\linewidth,height=0.75\linewidth]{"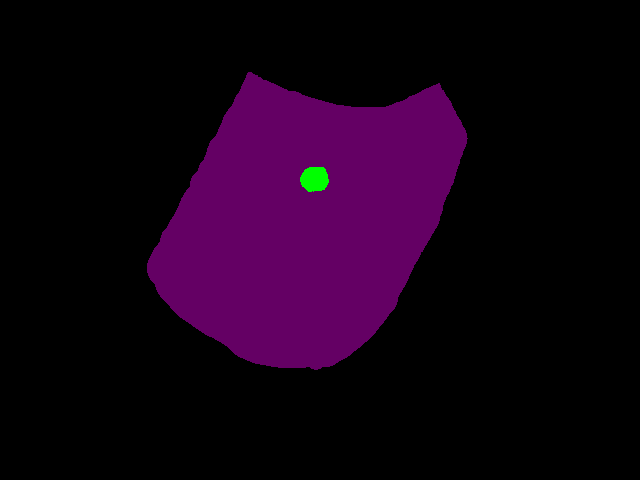"}
	\caption{Ground-truth}\label{fig:grt}
	\end{subfigure}
	\hfill
	
	\caption{Example of class imbalance problem}
	\label{fig:cls_imb}
	
\end{figure}

\begin{figure}

	\begin{subfigure}{0.425\linewidth}
	\includegraphics[width=\linewidth,height=0.75\linewidth]{"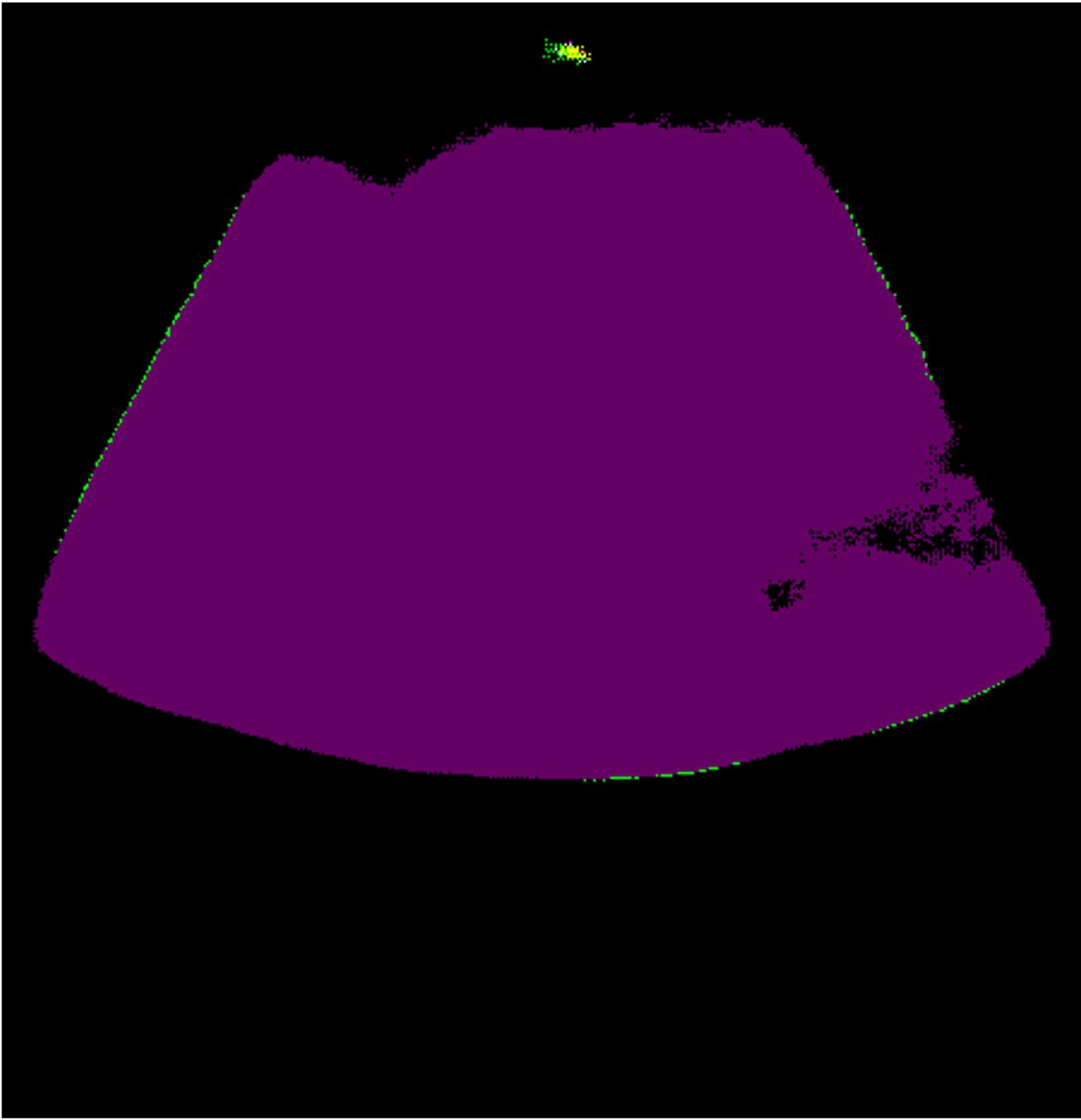"}
	\caption{Segmentation result}
	\end{subfigure}
	\hfill
	\begin{subfigure}{0.425\linewidth}
	\includegraphics[width=\linewidth,height=0.75\linewidth]{"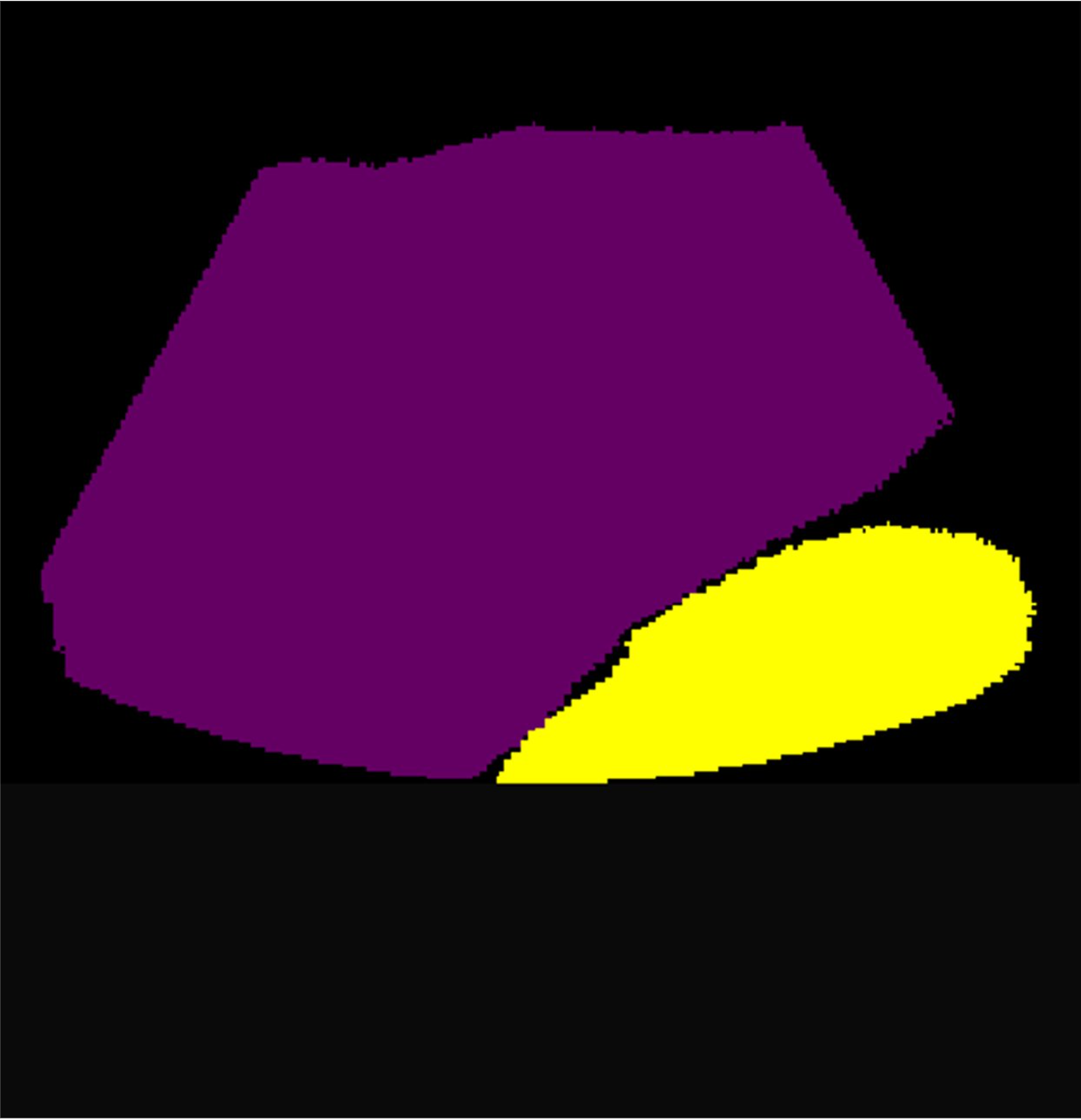"}
	\caption{Ground-truth}
	\end{subfigure}
	\hfill
	
	\caption{U-Net segmentation example}
	\label{fig:unet_res}
	
\end{figure}

In the abdominal section, different anatomical structures vary in shape and size, which may cause class imbalance for the traditional CNN segmentation algorithms like U-Net in \cite{Ronneberger2015UNetCN}. Although the total amount of instances may be almost equal in training, the relatively large organs and tissues occupy much more pixels in the ultrasound images. As shown in Fig. \ref{fig:cls_imb}, the violet part is the liver which occupies most of the pixels, and the green part is the gallbladder. This will make the algorithm classify as many pixels into the liver as possible, because a majority class has a much bigger influence on the final score than the minority class. Fig. \ref{fig:unet_res} shows the segmentation result of an ultrasound image containing liver (violet) and kidney (yellow). Compared with the ground truth, the result tends to ignore the kidney to focus on drawing the true mask of the liver. This example illustrates the necessity of introducing the FPN structure proposed by \cite{Lin2017FeaturePN}. 

\begin{figure}[ht]

	\centering
	\includegraphics[width=0.85\linewidth]{"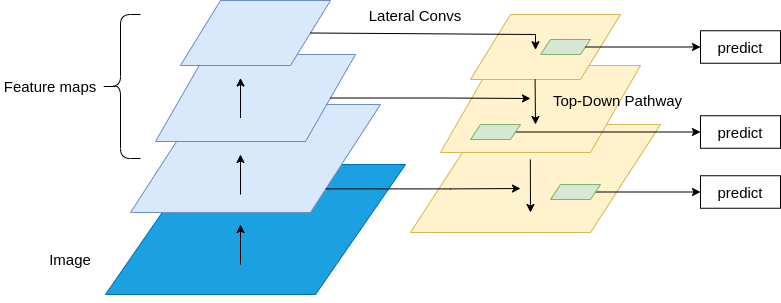"}
	\caption{Feature pyramid network structure}
	\label{fig:FPN}
	
\end{figure}

FPN takes the feature maps from multiple layers of the encoder rather than only from the deepest output. This pyramid network structure is scale-invariant in the sense that an object's scale changes with shifting its level in the feature pyramid. In other words, smaller objects are usually easier to be detected from smaller yet deeper feature maps, and vice versa. Compared with other pyramid network structures, FPN not only utilizes the relation between scale and layer depth, but also uses a top-down pathway to construct higher-resolution layers from a semantic layer. This solves the problem that feature maps composed of low-level structures (closer to the original level) are too naive for accurate object detection. As the reconstructed layers are semantically strong, but the locations of objects are not precise after all the down-sampling and up-sampling, the authors then added lateral connections between reconstructed layers and the corresponding feature maps to help the decoder predict the locations better. The overall structure of FPN is illustrated in Fig.~\ref{fig:FPN}.

\subsection{Spatial Recurrent Neural Network}
One important property of ultrasound images is that the anatomical structures form a constant spatial relationship under the same scan pattern. For example, in the midsagittal plane scanning, the mesenteric artery is usually at the bottom of the liver, and the pancreas is located at the side of the liver. Experienced sonographers rely heavily on such spatial context information to locate the target organs. This prior knowledge inspired us to take spatial context information into consideration.

To extract context information, the spatial recurrent neural network (SRNN) is introduced. Many studies have explored the utilization of RNNs to gather contextual information. \cite{650093} proposed a bidirectional recurrent neural network (BRNN) that passes both forward and backward across a time map to ensure the information is propagated across the entire timeline. When it comes to the context of spatial information, \cite{NIPS2008_66368270} proposed a multi-dimensional RNN to recognize handwriting. \cite{7298977} built a long short-term memory (LSTM) RNN structure for scene labeling. 

\begin{figure*}
	\centering
	\includegraphics[width=0.85\textwidth]{"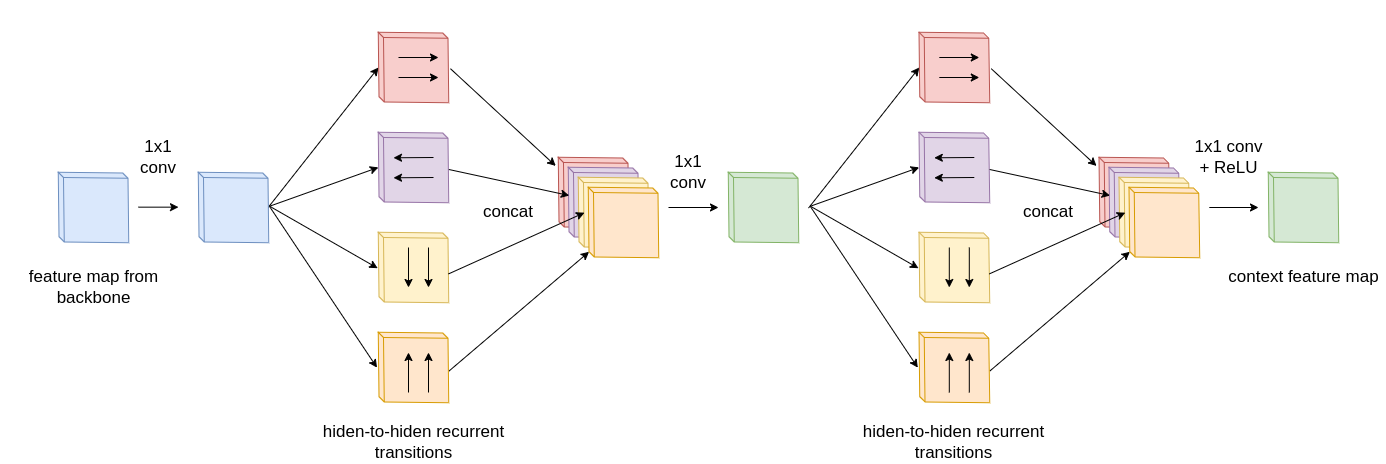"}
	\caption{Spatial RNN module}
	\label{fig:spatial_rnn}
\end{figure*}

\cite{Bell2016InsideOutsideND} proposed an object detection network structure called Inside-Outside Net (ION). Besides taking the information near an object's region of interest, the introduction of contextual information has improved the performance, for which a module of four directional RNNs is implemented. Fig. \ref{fig:spatial_rnn} shows the propagation of the RNNs. The structures are placed laterally across the feature maps, and move independently in four cardinal directions: right, left, down, up. The outputs from the RNNs are then concatenated and computed as a feature map containing both local and global contextual information.

\section{Multi Organ Segmentation Network}

\subsection{Network Structure Overview}
\begin{figure*}
	\centering
	\includegraphics[width=0.9\textwidth]{"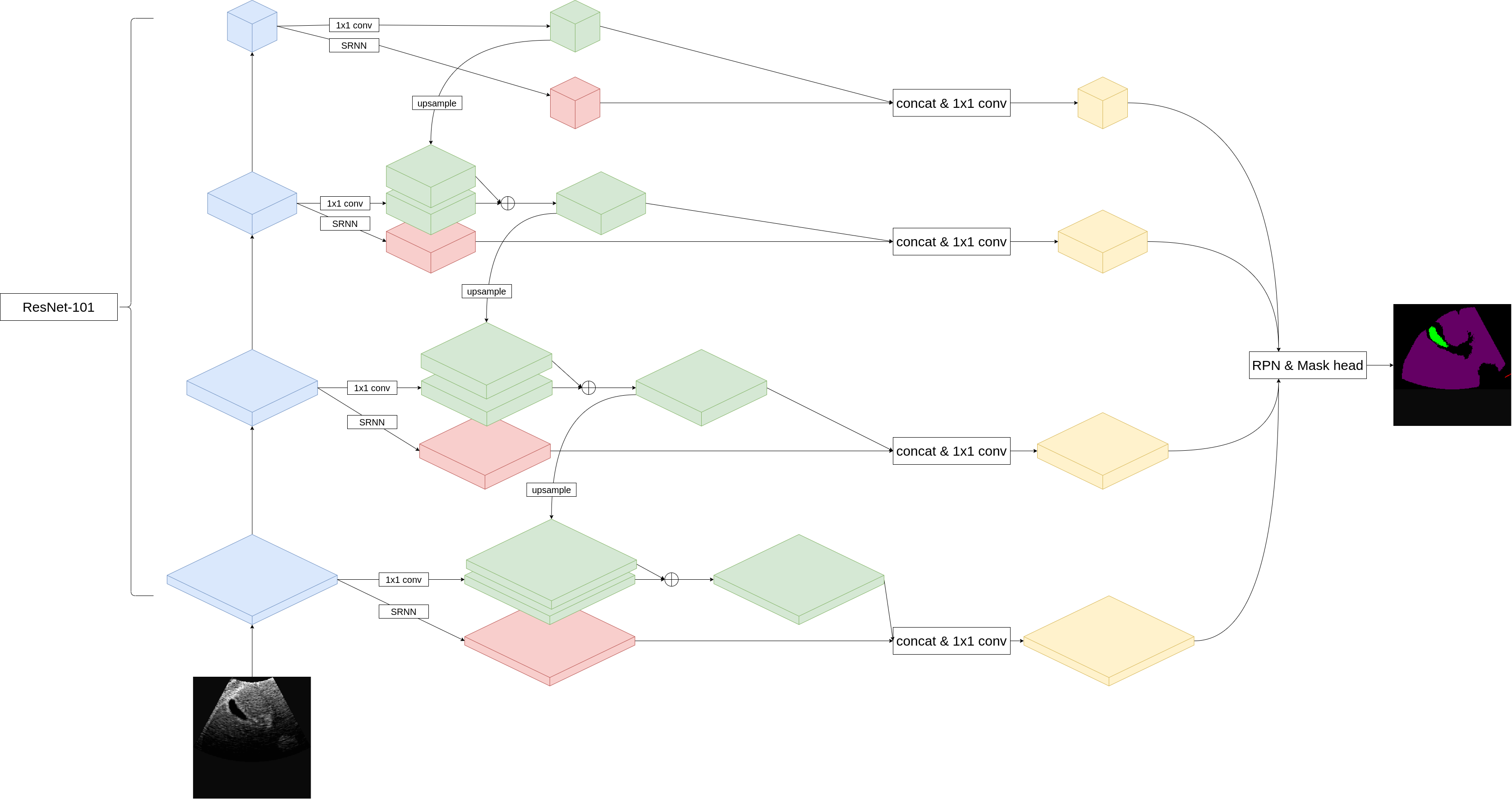"}
	\caption{Proposed network structure}
	\label{fig:cdefpn}
\end{figure*}

Fig.~\ref{fig:cdefpn} shows the structure of the proposed model. On the left side is the ResNet-101 backbone as the feature extractor. The input image is propagated from bottom to top, with the network generating feature maps of lower resolution and richer semantic information. We define the layers producing feature maps of the same size as one stage. We choose the output of the last layer of each stage to represent the output of the entire stage except the shallowest stage, because it is computationally time-consuming and of a low semantic feature. Each of the blue cubes represents an output of the stage called \{res2, res3, res4, res5\}, respectively. The feature maps go separately through a 1x1 convolution layer and the SRNN module.

The green cubes represent the feature maps after the convolution operation, and the red cubes represent the context feature maps. The deep feature map is concatenated with the context feature map, then it will be concatenated with the spatial context feature map. And the concatnated feature map will go through a normalization operation and be compressed to reduce depth channels. The sementic feature map from the upper layer, spatially coarser but semantically stronger, is upsampled by a scale factor of 2. 
Then the upsampled feature map from the upper pyramid level and the feature map from the current pyramid level are added together as the new feature map to be concatenated with the spatial feature map.

The yellow cubes are the final outputs of the entire feature extractor. After extracting semantic and spatial features, these pyramid feature maps are then sent to region proposal networks (RPN) and region-based detectors (Fast R-CNN). Unlike the classic object detectors, the FPN attaches RPN and Fast R-CNN to each of the output layers. The parameters of the heads are shared across all feature pyramid levels for simplicity, but the accuracy is actually very close with or without sharing parameters (\cite{Lin2017FeaturePN}). This is indirect proof that all the levels of the pyramid share similar semantic levels. After that, a DeepMask framework is used to generate masks. The structure of proposers and anchor/mask generators are omitted in the graph, since it is not our main interest.

\subsection{SRNN Structure}
\begin{figure}[ht]
	\centering
	\includegraphics[width=0.85\linewidth]{"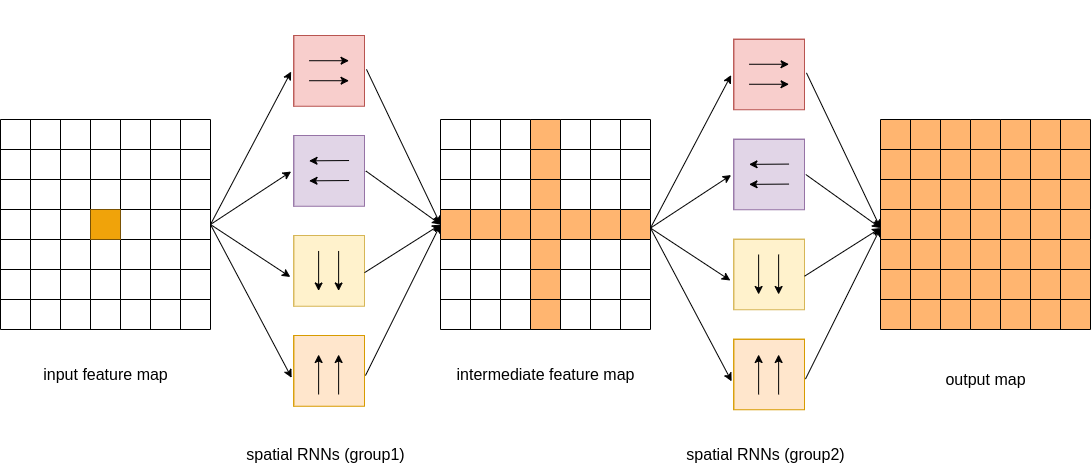"}
	\caption{Illustration of the IRNN propagation}
	\label{fig:irnn}
\end{figure}

The SRNN module follows the idea of the ION network structure. Fig.~\ref{fig:irnn} shows how the RNNs extract the contextual information. We first perform a 1x1 convolution to simulate the input-to-hidden data translation in the RNN. Then, four RNNs are propagated through the different directions mentioned above. The outputs from the RNNs are fused into an intermediate feature map. Until this step, each pixel contains the context information aiming at its four principal directions: right, left, up, down. Another round of the process is then repeated to further propagate the aggregated spatial context information in each principal direction. Finally, a feature map containing the overall context information is generated. For comparison, in the feature map on the left in Fig.~\ref{fig:irnn}, each pixel only contains information about itself and its neighbors (depending on the perspective field). After the first round of RNN propagation, the pixels get the context information from its 4 directions. Finally, RNNs propagate through the context-rich pixels to extract the full-directional context information. Therefore, the last feature map is globally context-rich.

\subsection{IRNN}
An RNN is specialized for processing sequential data. 
The data fed into the input nodes is propagated through the hidden nodes, updating the internal states using past and present data. 
There are variants of RNN such as gated recurrent units (\cite{Cho2014OnTP}), LSTM (\cite{lstm}), and plain tanh RNNs. The RNN in this work follows the model in \cite{Le2015ASW} due to its efficiency and simplicity of training. This RNN structure is called IRNN, because the recurrent weight matrix is initialized to the identity matrix. IRNN has a good performance for long-range data dependencies (\cite{Bell2016InsideOutsideND}). IRNN is composed of the rectified linear unit, and the recurrent weight matrix is initialized to the identity matrix. Therefore, gradients are propagated backward with full strength at initialization. We adapt four independent IRNNs that propagate through four different directions. Given below is the update function for the IRNN moving from left to right. The rest IRNNs follow a similar equation according to the propagation direction:
\begin{equation} \label{hidden}
{h}_{i,j}^{right}\leftarrow max(W_{hh}^{right}h_{i,j-1}^{right}+h_{i,j}^{right},0),
\end{equation}
where $W$ is the hidden transition matrix and $h_{i,j}$ is the feature at pixel$(i,j)$.

Each direction on independent rows or columns is computed in parallel, and the output from the IRNN is computed by concatenating the hidden state from the four directions at each spatial location.

\section{Experiments}

\subsection{Dataset}
\begin{figure}

	\begin{subfigure}{0.425\linewidth}
	\includegraphics[width=\linewidth,height=0.75\linewidth]{"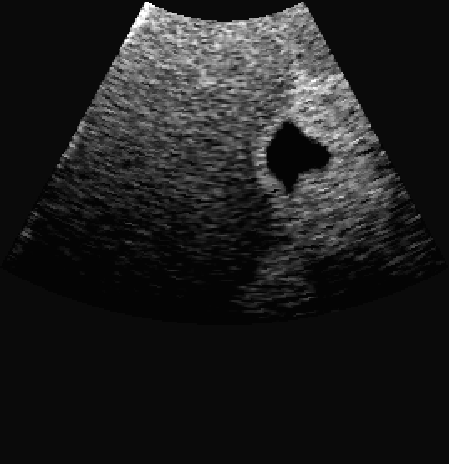"}
	\caption{CT-generated image}\label{fig:a_ultimage}
	\end{subfigure}
	\hfill
	\begin{subfigure}{0.425\linewidth}
	\includegraphics[width=\linewidth,height=0.75\linewidth]{"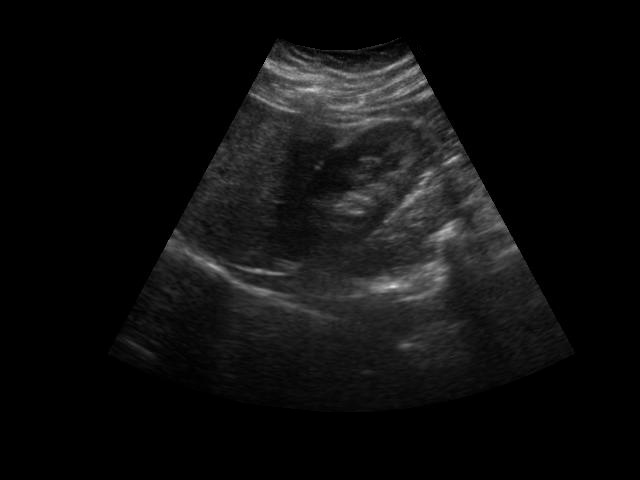"}
	\caption{Regular image}\label{fig:r_ultimahe}
	\end{subfigure}
	\hfill
	
	\caption{Example of images in dataset}
	\label{fig:dataset}
	
\end{figure}

A dataset of high quality is one of the key factors to train a neural network. Unfortunately, there are few open-source abdominal ultrasound image datasets. Most of the relevant researchers have not made their dataset public for the protection of patients' privacy. In this work, we use the dataset provided by \cite{dataset} containing both artificial ultrasound images which are translated from CT images, and a few images generated from real ultrasound scans. In the work of \cite{dataset}, they applied generative neural networks trained with a cycle consistency loss, and successfully improved the realism in ultrasound simulation from computed tomography (CT). We use 926 labeled artificial ultrasound images and 61 labeled real ultrasound scans, in which we can have the annotations of the liver, kidney, gallbladder, spleen, and vessels. Different organs are assigned segmentation masks of different colors. Table \ref{tb:dataset} shows the name of the anatomical structures and the corresponding instance number. We mixed and separated the dataset into 3 subsets: 787 images for training, 100 images for testing, and 100 images for validation. Since there is a huge difference in the image quality between CT-generated ultrasound images and real ultrasound images(see Fig.~\ref{fig:dataset}), we believe the performance of the proposed model can still be improved if we can have access to high-quality ultrasound image datasets.

\begin{table}[hb]
\begin{center}
\caption{Dataset}\label{tb:dataset}
\begin{tabular}{|c|c|c|c|c|c|}
\hline
Name & Liver & Kidney & Gallbladder & Vessels & Spleen\\\hline
Number & 591 & 377 & 219 & 289 & 172 \\\hline
Color & violet & yellow & green & red & pink \\\hline
\end{tabular}
\end{center}
\end{table}

\subsection{Detectron2}
\cite{wu2019detectron2} from FaceBook research team released a powerful object detection tool called \emph{detectron2} containing many network architectures and training tools. We build the backbone framework based on the implementation of FPN in \emph{detectron2}. And we develop our SRNN structure inserted into the FPN framework as a new feature extractor. The standardized region proposal network(RPN), Fast R-CNN, and Mask R-CNN heads are attached after the feature extractor as the proposal generators. Specifically, the output feature maps are from \{res2, res3, res4, res5\} of the ResNet layers. The size of the anchor generators are set to $32\times 32, 64\times 64, 128\times 128$, and $256 \times  256$. For each feature map, FPN gives 1000 proposals. The region of interests(ROI) box head follows the structure of Fast R-CNN with 2 fully convolutional layers and $7\times7$ pooler resolution. The Mask R-CNN head has 4 convolutional layers and a pooler resolution of $14\times14$. The ROI heads score threshold is set to 0.5 for both box and mask heads. We have made some modifications to the model, enabling it to run under the \emph{detectron2} framework. For example, we use a small learning rate to ensure there will not be NaN and infinity scores in the final result and reduce the ROI head batch size from 512 to 128, which is computationally efficient while the accuracy is nearly the same.

\subsection{Loss Functions}
Multiple loss functions are included in our training procedure, some of which are listed here:

\subsubsection{Objectness loss}

For detection of the object appearance, the binary cross-entropy loss is used in the RPN head. This loss is only for the classification of object and background. In RPN, this loss will be computed on the objectness logits feature map and the ground truth objectness logits. If the pixel is from some target object, it will be marked as ``1'', otherwise ``0''. The formulation is:
\begin{equation}
    L_{obj}=-\frac{1}{n}\sum_{i=1}^{n}y_i \log{(p(y_i))}+(1-y_i) \log{(1-p(y_i))},
\end{equation}
where $y$ is the label (``1'' for foreground, ``0'' for background), and $p(y)$ is the predicted probability of instance existence for all the grid points in the feature map.

\subsubsection{Anchor and bounding box loss}

Both RPN and ROI(Box) heads use a smooth l1 loss for the proposed anchors and bounding boxes. The anchors and bounding boxes are represented as a tensor of length 4: $(x,y,w,h)$, namely the $x,y$ coordinates and width/height of the anchor or bounding box. Then with the ground truth information, 4 deltas ($d_x$, $d_y$, $d_w$, $d_h$) are calculated by
\begin{equation}
    \begin{split}
    d_x & =(g_x-p_x)\\
    d_y & =(g_y-p_y)\\
    d_w & =\log{(g_w/p_w)}\\
    d_h & =\log{(g_h/p_g)}
    \end{split}
\end{equation}
where $g$ represents the ground truth and $p$ stand for the predicted anchor or bounding box. The deltas will be stacked together to compute the smooth l1 loss, given by
\begin{equation}
    L_{1}^{smooth}(x)=
    \begin{cases}
    0.5x^2   & if \quad |x|<\beta\\
    |x|-0.5*\beta   & otherwise
    \end{cases}
\end{equation}
where $\beta$ is a pre-defined smooth parameter.

\subsubsection{Classification loss}

Softmax cross entropy loss is calculated for all the foreground and background prediction scores:
\begin{equation}
    L_{CE}=-\sum_{i=1}^{n} y_i log(p_i)
\end{equation}
where $y_i$ is the true label and $p_i$ is the softmax probability for the $i^{th}$ class.\\

\subsubsection{Mask loss}

The mask loss is defined as the average binary cross-entropy loss. Eq. \ref{eq:loss_mask} computes the mask loss for the $k^{th}$ class:
\begin{equation}
    L_{mask}=-\frac{1}{m^2}\sum_{1\leq i,j\leq m}[y_{i,j} \log \Hat{y}_{i,j}^{k}+(1-y_{i,j})\log (1-\Hat{y}_{i,j}^{k})]
    \label{eq:loss_mask}
\end{equation}
where $y_{i,j}$ is the label of a cell($i$,$j$) in the true mask for the region of size $m\times m$, and $\Hat{y}_{i,j}^k$ represents the value of the same cell in the predicted mask.

\subsection{Experiment Setup}

\begin{figure}
	\begin{subfigure}{0.4\linewidth}
	\includegraphics[width=\linewidth]{"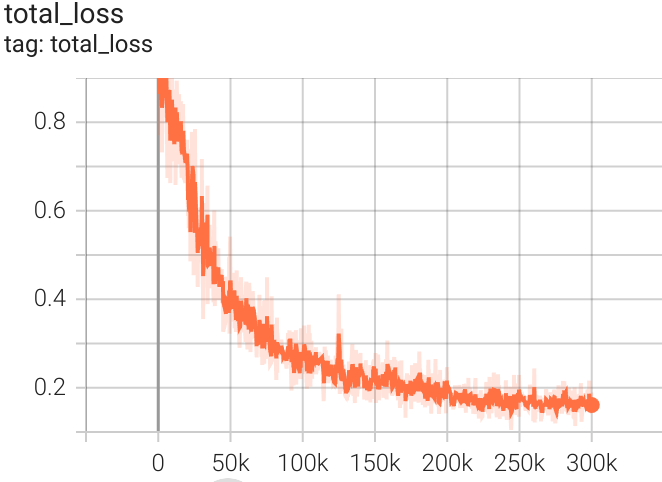"}
	\caption{Total}\label{fig:total_loss}
	\end{subfigure}
	\hfill
	\begin{subfigure}{0.4\linewidth}
	\includegraphics[width=\linewidth]{"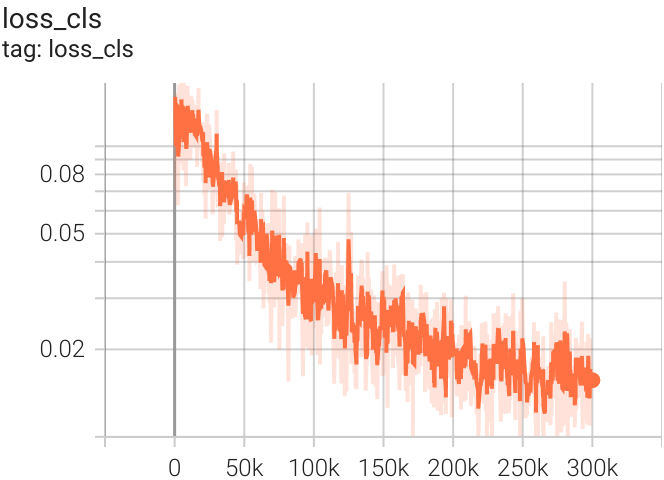"}
	\caption{Class}\label{fig:loss_cls}
	\end{subfigure}
	
	\begin{subfigure}{0.4\linewidth}
	\includegraphics[width=\linewidth]{"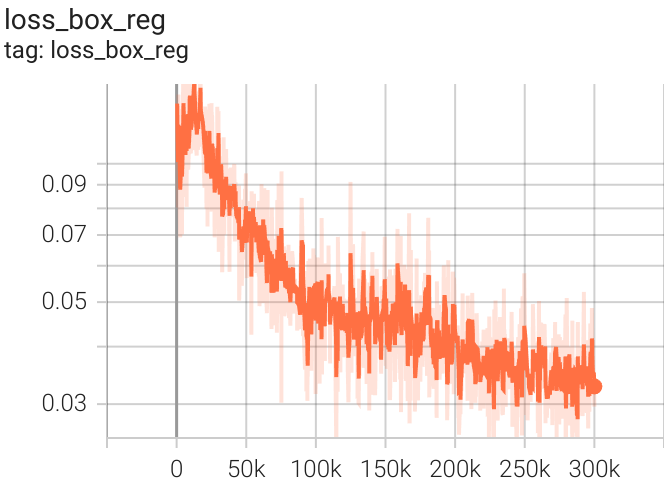"}
	\caption{Bbox}\label{fig:loss_box_reg}
	\end{subfigure}
	\hfill
	\begin{subfigure}{0.4\linewidth}
	\includegraphics[width=\linewidth]{"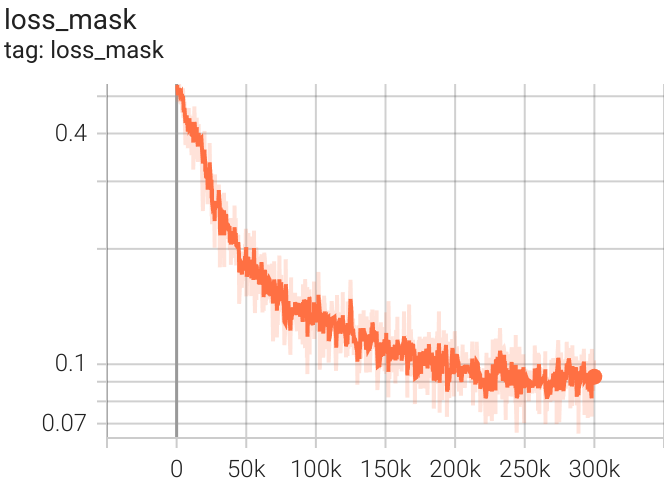"}
	\caption{Mask}\label{fig:loss_mask}
	\end{subfigure}
	
	\caption{Loss curves}
	\label{fig:loss_curve}
\end{figure}

Our experiment builds upon \emph{detectron2} framework in PyTorch environment. We modified the original FPN in \emph{detectron2} via adding the SRNN module. 
We train the model on a single GPU (RTX3080). The batch size is set to 1 since this GPU has relatively limited memory and the dataset is again relatively small. The learning rate is set to 0.0025. The model is trained for 300k epochs, taking around 30 hours to converge. We also tried to explore deeper layers in the backbone, adding 2 extra pyramid layers on the top of the backbone, but the accuracy fails to increase. Fig.~\ref{fig:loss_curve} shows the converge curve of the proposed model, namely the loss curve of (a) total loss, (b) classification loss, (c) bounding box region loss, (d) and mask loss.

\section{Results}

\subsection{Quantitative Result}

The performance of the trained model is evaluated by the following dice coefficient:
\begin{equation}
    D = \frac{2|X\cap Y|}{|X|+|Y|}
\end{equation}

Dice coefficient\cite{sorensen1948method} is one of the most widely used evaluation methods in the research field of image segmentation. We use dice coefficient as our evaluation metric for the convenience of comparison with other research. The dice coefficient is twice the number of elements common to two sets $X$ and $Y$, divided by the sum of the number of elements in each set. In our work, $X$ and $Y$ are the predicted classification map and the ground truth. Therefore, the numerator is regarded as the intersection pixels of the predicted mask and the ground truth, and the denominator the sum of mask pixels in both. Considering we have 5 object classes (background not included), the coefficient score is computed separately and then averaged as the final score. As there might be no appearance of certain classes, we added a smoothing parameter $\epsilon$ to avoid zeros in the denominator. This smoothing parameter can be arbitrarily small, and we set it to $1\times10^{-6}$. The influence on the evaluation result from the smoothing parameter can be ignored as long as it is smaller than the given setting. The modified equation is given in~\eqref{eq:modDice}, where $n$ is the number of classes:
\begin{equation}
    D = \frac{\sum_{i=1}^{n}\frac{2|X_i\cap Y_i|}{|X_i|+|Y_i|+\epsilon}}{n}
    \label{eq:modDice}
\end{equation}

There are few similar researches surrounding abdominal multi-organ segmentation. Neither any relevant benchmark nor competition exists. Therefore, we separately pick some comparable results from different researches aiming at single organ segmentation. Respectively, the segmentation result of liver is compared with the work of \cite{9957634}. \cite{7318778} proposed a segmentation network targeting at kidney. Their result is taken into comparison as well. And the segmentation performance of gallbladder and spleen are compared with the work of \cite{gallbladder} and \cite{YUAN2022103724}. The segmentation performance of vessels is not compared with other works, because most of the relevant research are focused on the segmentation of cardiac arteries. The numerical result may not seem encouraging compared with those well-aimed researches. On one hand, the segmentation performance is limited by our lack of high-quality data. In our research, we have only 172 instances of training spleen samples, not to mention that most of the ultrasound images are pseudo ultrasound images interpreted from CT image. On the other hand, the specifically targeted researches usually introduced some prior knowledge into their segmentation algorithm like the detection of boundaries. Meanwhile, we trained a pure FPN model for comparison to demonstrate the improvement brought by SRNN. Table~\ref{tb:result} shows the dice score of each class, where we can see that the improvement of performance by SRNN is significant. The proposed model outperformed the pure FPN model.

\begin{table}[hb]
\begin{center}
\caption{Evaluation Result}\label{tb:result}
\begin{tabular}{|c|c|c|c|}
\hline
Organ/Tissue &Related Work & FPN & FPN+SRNN \\\hline
Liver  & 0.821 by Man et al. & 0.907 & 0.924\\\hline
Kidney  & 0.5 by Marsousi et al. & 0.806 & 0.836\\\hline
Gallbladder  & 0.893 by Lian et al. & 0.799 & 0.815\\\hline
Vessels  & -- & 0.801 & 0.825\\\hline
Spleen  & 0.93 by Yuan et al. & 0.810 & 0.859\\\hline
Average  & -- & 0.840 & 0.865\\\hline

\end{tabular}
\end{center}
\end{table}

\subsection{Qualitative Result}

\begin{figure}[ht]
	\begin{subfigure}{0.4\linewidth}
	\includegraphics[width=\linewidth]{"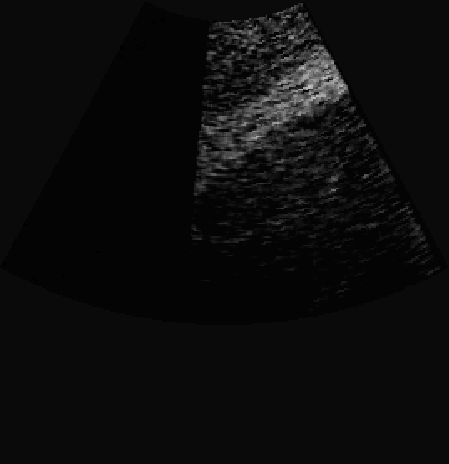"}
	\caption{Original image}\label{fig:ori}
	\end{subfigure}
	\hfill
	\begin{subfigure}{0.4\linewidth}
	\includegraphics[width=\linewidth]{"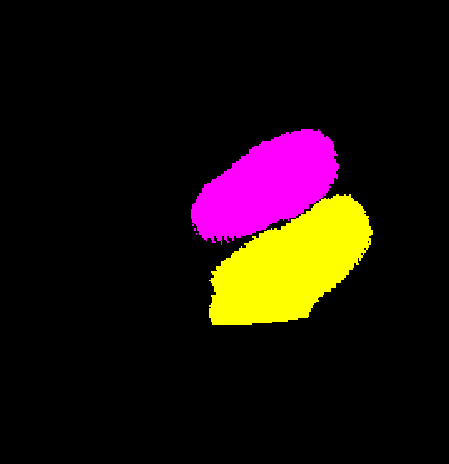"}
	\caption{Ground truth}\label{fig:g_t}
	\end{subfigure}
	
	\begin{subfigure}{0.4\linewidth}
	\includegraphics[width=\linewidth]{"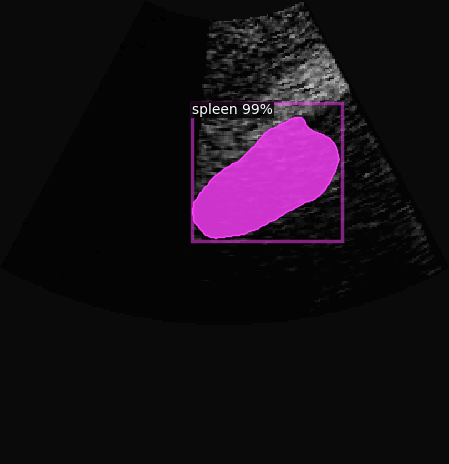"}
	\caption{FPN result}\label{fig:lfpn_res}
	\end{subfigure}
	\hfill
	\begin{subfigure}{0.4\linewidth}
	\includegraphics[width=\linewidth]{"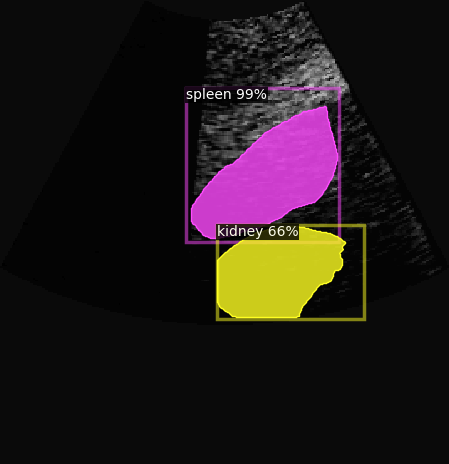"}
	\caption{Proposed model result}\label{fig:fpns_res}
	\end{subfigure}
	\caption{Test results}
	\label{fig:test_res}
\end{figure}

We have tested the proposed model on the artificial and real ultrasound images from the evaluation data. Fig.~\ref{fig:test_res} shows an example of the semantic segmentation on ultrasound image. (a) is the original ultrasound image, (b) is the corresponding ground truth. (c) and (d) are the segmentation result generated by the pure FPN and our proposed model. We can see that the proposed model outperforms the pure FPN.

\begin{figure}[ht]
	\begin{subfigure}{0.4\linewidth}
	\includegraphics[width=\linewidth]{"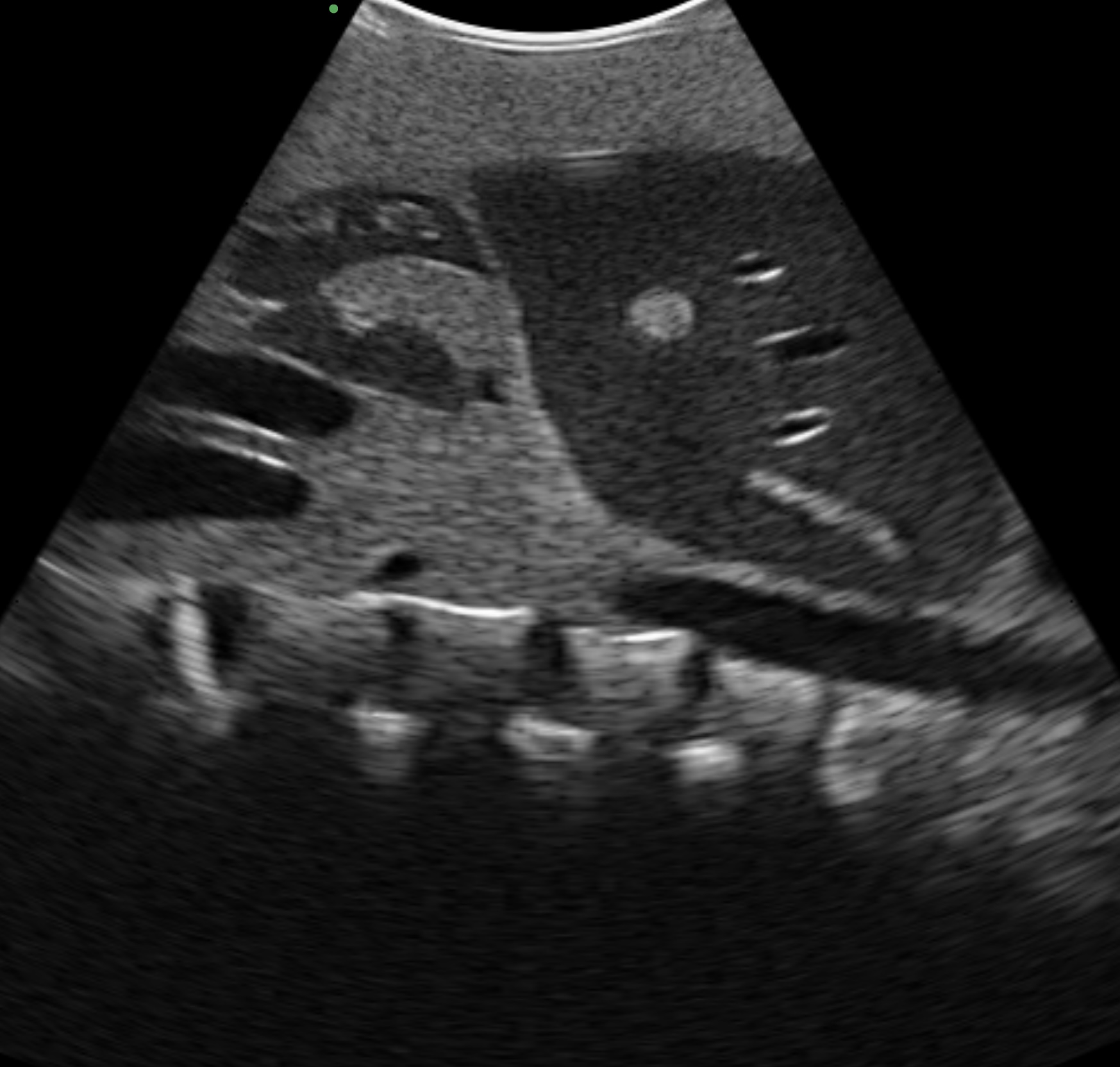"}
	\caption{In vitro image}\label{fig:ri}
	\end{subfigure}
	\hfill
	\begin{subfigure}{0.4\linewidth}
	\includegraphics[width=\linewidth]{"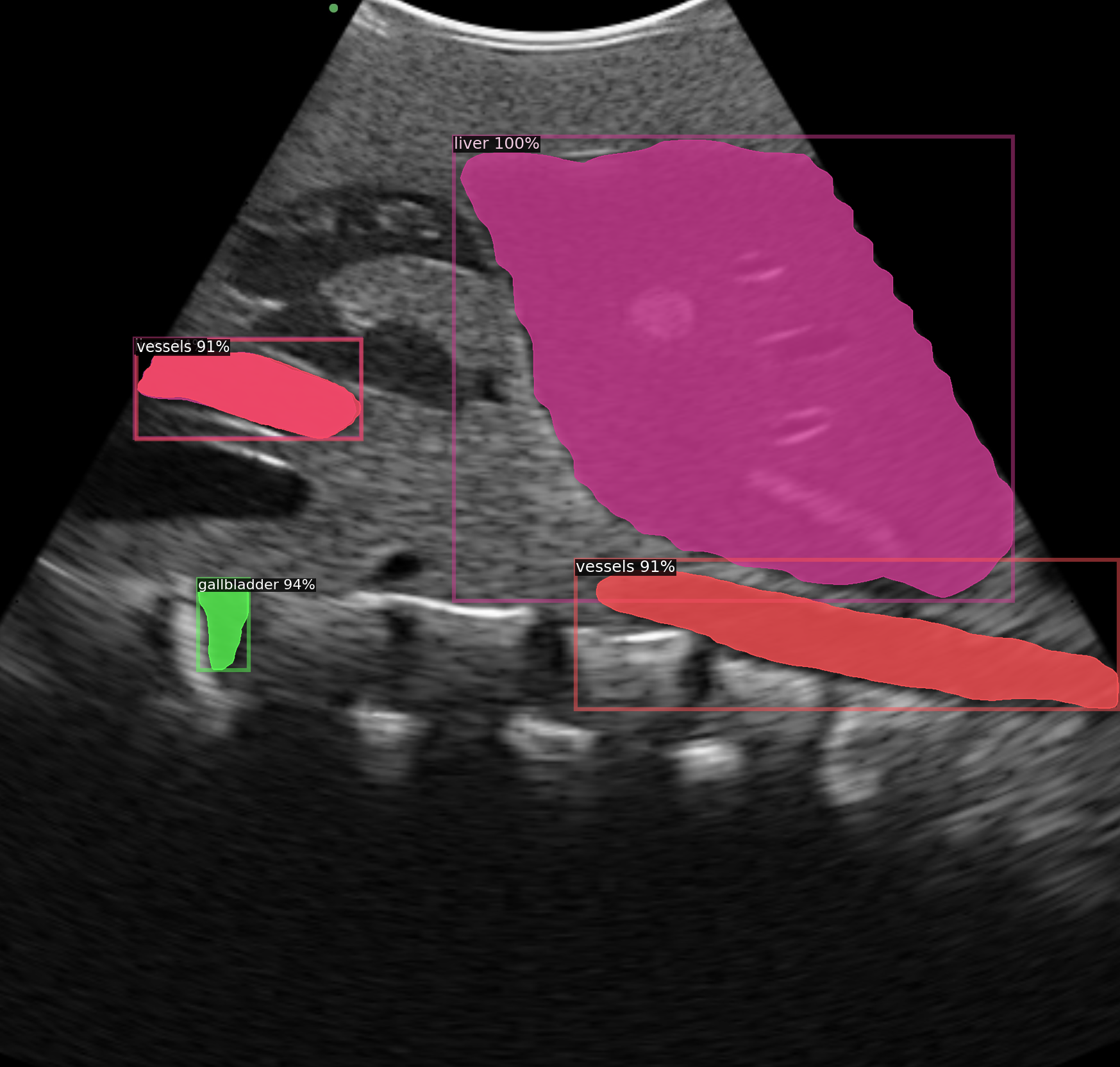"}
	\caption{Proposed model result}\label{fig:ri_res}
	\end{subfigure}
	\hfill
	\caption{Robot-assisted ultrasound image capture/segmentation}
	\label{fig:phantom_image}
\end{figure}

Furthermore, our proposed model was tested on the ultrasound images collected manually from an abdominal phantom in our laboratory. Fig.~\ref{fig:phantom_image} shows an outstanding performance of our model: (a) is an ultrasound image collected from the phantom, (b) is the semantic masks generated.

\section{Conclusions and Future Work}

We proposed an FPN based multi-organ/tissue segmentation method combined with the utilization of SRNN. From the experimental results, we can see that the introduction of spatial context information has improved the performance of the original FPN model both in quantitative and qualitative comparison. The findings of this work would benefit from further research including different scan patterns, since a prior knowledge of the ultrasound scan pattern would help add more precise spatial context information.
  
  

\bibliography{ifacconf}             

\begin{thebibliography}{27}
\providecommand{\natexlab}[1]{#1}
\providecommand{\url}[1]{\texttt{#1}}
\providecommand{\urlprefix}{URL }
\expandafter\ifx\csname urlstyle\endcsname\relax
  \providecommand{\doi}[1]{doi:\discretionary{}{}{}#1}\else
  \providecommand{\doi}{doi:\discretionary{}{}{}\begingroup
  \urlstyle{rm}\Url}\fi

\bibitem[{Almajalid et~al.(2018)Almajalid, Shan, Du, and Zhang}]{8614204}
Almajalid, R., Shan, J., Du, Y., and Zhang, M. (2018).
\newblock Development of a deep-learning-based method for breast ultrasound
  image segmentation.
\newblock In \emph{IEEE International Conference on Machine Learning and
  Applications}, 1103--1108.

\bibitem[{Bell et~al.(2016)Bell, Zitnick, Bala, and
  Girshick}]{Bell2016InsideOutsideND}
Bell, S., Zitnick, C.L., Bala, K., and Girshick, R.B. (2016).
\newblock Inside-outside net: Detecting objects in context with skip pooling
  and recurrent neural networks.
\newblock \emph{IEEE Conference on Computer Vision and Pattern Recognition},
  2874--2883.

\bibitem[{Boukerroui et~al.(2003)Boukerroui, Baskurt, Noble, and
  Basset}]{BOUKERROUI2003779}
Boukerroui, D., Baskurt, A., Noble, J., and Basset, O. (2003).
\newblock Segmentation of ultrasound images––multiresolution 2d and 3d
  algorithm based on global and local statistics.
\newblock \emph{Pattern Recognition Letters}, 24(4), 779--790.

\bibitem[{Byeon et~al.(2015)Byeon, Breuel, Raue, and Liwicki}]{7298977}
Byeon, W., Breuel, T.M., Raue, F., and Liwicki, M. (2015).
\newblock Scene labeling with lstm recurrent neural networks.
\newblock In \emph{IEEE Conference on Computer Vision and Pattern Recognition},
  3547--3555.

\bibitem[{Chen et~al.(2022)Chen, Yin, Dai, Zhang, Yin, and
  Cui}]{CHEN2022106712}
Chen, G., Yin, J., Dai, Y., Zhang, J., Yin, X., and Cui, L. (2022).
\newblock A novel convolutional neural network for kidney ultrasound images
  segmentation.
\newblock \emph{Computer Methods and Programs in Biomedicine}, 218, 106712.

\bibitem[{Cho et~al.(2014)Cho, van Merrienboer, Bahdanau, and
  Bengio}]{Cho2014OnTP}
Cho, K., van Merrienboer, B., Bahdanau, D., and Bengio, Y. (2014).
\newblock On the properties of neural machine translation: Encoder–decoder
  approaches.
\newblock In \emph{SSST@EMNLP}.

\bibitem[{Graves and Schmidhuber(2008)}]{NIPS2008_66368270}
Graves, A. and Schmidhuber, J. (2008).
\newblock Offline handwriting recognition with multidimensional recurrent
  neural networks.
\newblock In D.~Koller, D.~Schuurmans, Y.~Bengio, and L.~Bottou (eds.),
  \emph{Advances in Neural Information Processing Systems}, volume~21. Curran
  Associates, Inc.

\bibitem[{He et~al.(2016)He, Zhang, Ren, and Sun}]{He2016DeepRL}
He, K., Zhang, X., Ren, S., and Sun, J. (2016).
\newblock Deep residual learning for image recognition.
\newblock \emph{IEEE Conference on Computer Vision and Pattern Recognition},
  770--778.

\bibitem[{Hochreiter and Schmidhuber(1997)}]{lstm}
Hochreiter, S. and Schmidhuber, J. (1997).
\newblock Long short-term memory.
\newblock \emph{Neural computation}, 9, 1735--80.

\bibitem[{Huang et~al.(2021)Huang, Chen, Xu, Chen, Yu, Cai, and
  Zhang}]{crossorgan}
Huang, H., Chen, H., Xu, H., Chen, Y., Yu, Q., Cai, Y., and Zhang, Q. (2021).
\newblock Cross-tissue/organ transfer learning for the segmentation of
  ultrasound images using deep residual u-net.
\newblock \emph{Journal of Medical and Biological Engineering}, 41.
\newblock \doi{10.1007/s40846-020-00585-w}.

\bibitem[{Huang et~al.(2019)Huang, Xue, and Wu}]{10.1371/journal.pone.0219369}
Huang, W., Xue, Y., and Wu, Y. (2019).
\newblock A cad system for pulmonary nodule prediction based on deep
  three-dimensional convolutional neural networks and ensemble learning.
\newblock \emph{PLOS ONE}, 14(7), 1--17.

\bibitem[{Le et~al.(2015)Le, Jaitly, and Hinton}]{Le2015ASW}
Le, Q.V., Jaitly, N., and Hinton, G.E. (2015).
\newblock A simple way to initialize recurrent networks of rectified linear
  units.
\newblock \emph{ArXiv}, abs/1504.00941.

\bibitem[{Lei et~al.(2021)Lei, Wang, Roper, Jani, Patel, Curran, Patel, Liu,
  and Yang}]{malepelvic}
Lei, Y., Wang, T., Roper, J., Jani, A., Patel, S., Curran, W., Patel, P., Liu,
  T., and Yang, X. (2021).
\newblock Male pelvic multi‐organ segmentation on transrectal ultrasound
  using anchor‐free mask cnn.
\newblock \emph{Medical Physics}, 48.
\newblock \doi{10.1002/mp.14895}.

\bibitem[{Lian et~al.(2017)Lian, Ma, ma, Shi, Liu, Yang, and Guo}]{gallbladder}
Lian, J., Ma, Y., ma, Y., Shi, B., Liu, J., Yang, Z., and Guo, Y. (2017).
\newblock Automatic gallbladder and gallstone regions segmentation in
  ultrasound image.
\newblock \emph{International Journal of Computer Assisted Radiology and
  Surgery}, 12.
\newblock \doi{10.1007/s11548-016-1515-z}.

\bibitem[{Lin et~al.(2017)Lin, Doll{\'a}r, Girshick, He, Hariharan, and
  Belongie}]{Lin2017FeaturePN}
Lin, T.Y., Doll{\'a}r, P., Girshick, R.B., He, K., Hariharan, B., and Belongie,
  S.J. (2017).
\newblock Feature pyramid networks for object detection.
\newblock \emph{IEEE Conference on Computer Vision and Pattern Recognition},
  936--944.

\bibitem[{Man et~al.(2022)Man, Wu, Man, Shi, Wang, and Liang}]{9957634}
Man, L., Wu, H., Man, J., Shi, X., Wang, H., and Liang, Q. (2022).
\newblock Machine learning for liver and tumor segmentation in ultrasound based
  on labeled ct and mri images.
\newblock In \emph{2022 IEEE International Ultrasonics Symposium (IUS)}, 1--4.
\newblock \doi{10.1109/IUS54386.2022.9957634}.

\bibitem[{Marsousi et~al.(2015)Marsousi, Plataniotis, and
  Stergiopoulos}]{7318778}
Marsousi, M., Plataniotis, K.N., and Stergiopoulos, S. (2015).
\newblock Atlas-based segmentation of abdominal organs in 3d ultrasound, and
  its application in automated kidney segmentation.
\newblock In \emph{2015 37th Annual International Conference of the IEEE
  Engineering in Medicine and Biology Society (EMBC)}, 2001--2005.
\newblock \doi{10.1109/EMBC.2015.7318778}.

\bibitem[{Mignotte and Meunier(2001)}]{MIGNOTTE2001265}
Mignotte, M. and Meunier, J. (2001).
\newblock A multiscale optimization approach for the dynamic contour-based
  boundary detection issue.
\newblock \emph{Computerized Medical Imaging and Graphics}, 25(3), 265--275.

\bibitem[{Mignotte et~al.(2001)Mignotte, Meunier, and Tardif}]{Mignotte2001}
Mignotte, M., Meunier, J., and Tardif, J.C. (2001).
\newblock Endocardial boundary e timation and tracking in echocardiographic
  images using deformable template and markov random fields.
\newblock \emph{Pattern Anal. Appl.}, 4, 256--271.

\bibitem[{Mishra et~al.(2003)Mishra, Dutta, and Ghosh}]{MISHRA2003967}
Mishra, A., Dutta, P., and Ghosh, M. (2003).
\newblock A ga based approach for boundary detection of left ventricle with
  echocardiographic image sequences.
\newblock \emph{Image and Vision Computing}, 21(11), 967--976.

\bibitem[{Ronneberger et~al.(2015)Ronneberger, Fischer, and
  Brox}]{Ronneberger2015UNetCN}
Ronneberger, O., Fischer, P., and Brox, T. (2015).
\newblock U-net: Convolutional networks for biomedical image segmentation.
\newblock \emph{ArXiv}, abs/1505.04597.

\bibitem[{Schuster and Paliwal(1997)}]{650093}
Schuster, M. and Paliwal, K. (1997).
\newblock Bidirectional recurrent neural networks.
\newblock \emph{IEEE Transactions on Signal Processing}, 45(11), 2673--2681.

\bibitem[{Sorensen(1948)}]{sorensen1948method}
Sorensen, T.A. (1948).
\newblock A method of establishing groups of equal amplitude in plant sociology
  based on similarity of species content and its application to analyses of the
  vegetation on danish commons.
\newblock \emph{Biol. Skar.}, 5, 1--34.

\bibitem[{Vitale et~al.(2019)Vitale, Orlando, Iarussi, and Larrabide}]{dataset}
Vitale, S., Orlando, J., Iarussi, E., and Larrabide, I. (2019).
\newblock Improving realism in patient-specific abdominal ultrasound simulation
  using cyclegans.
\newblock \emph{International Journal of Computer Assisted Radiology and
  Surgery}.
\newblock \doi{10.1007/s11548-019-02046-5}.

\bibitem[{Wu et~al.(2019)Wu, Kirillov, Massa, Lo, and
  Girshick}]{wu2019detectron2}
Wu, Y., Kirillov, A., Massa, F., Lo, W.Y., and Girshick, R. (2019).
\newblock Detectron2.
\newblock \url{https://github.com/facebookresearch/detectron2}.

\bibitem[{Yuan et~al.(2022)Yuan, Puyol-Antón, Jogeesvaran, Smith, Inusa, and
  King}]{YUAN2022103724}
Yuan, Z., Puyol-Antón, E., Jogeesvaran, H., Smith, N., Inusa, B., and King,
  A.P. (2022).
\newblock Deep learning-based quality-controlled spleen assessment from
  ultrasound images.
\newblock \emph{Biomedical Signal Processing and Control}, 76, 103724.
\newblock \doi{https://doi.org/10.1016/j.bspc.2022.103724}.

\bibitem[{Zhang et~al.(2016)Zhang, Ying, Yang, Ahuja, and Chen}]{7822557}
Zhang, Y., Ying, M.T.C., Yang, L., Ahuja, A.T., and Chen, D.Z. (2016).
\newblock Coarse-to-fine stacked fully convolutional nets for lymph node
  segmentation in ultrasound images.
\newblock In \emph{IEEE International Conference on Bioinformatics and
  Biomedicine}, 443--448.

\end{thebibliography}

\end{document}